\documentclass{article}

\PassOptionsToPackage{numbers, compress}{natbib}

\usepackage[final]{neurips_2024}

\usepackage[utf8]{inputenc} %
\usepackage[T1]{fontenc}    %
\usepackage{url}            %
\usepackage{booktabs}       %
\usepackage{amsfonts}       %
\usepackage{nicefrac}       %
\usepackage{microtype}      %
\usepackage{xcolor}         %
\usepackage{graphicx}
\usepackage{booktabs}
\usepackage{multirow}
\usepackage{enumitem}
\usepackage{subcaption}
\usepackage{caption}
\usepackage{tikz}

\usepackage[pdftex,bookmarksnumbered,bookmarksopen,colorlinks,citecolor={blue!80!black},linkcolor={blue!80!black},urlcolor={blue!80!black}]{hyperref}

\newcommand{\ours}{\textsc{crate}-$\alpha$}

\usepackage{bm}
\usepackage{amsmath}
\newcommand{\vX}{\bm{X}}
\newcommand{\vx}{\bm{x}}

\newcommand{\R}{\mathbb{R}}

\newcommand{\vA}{\bm{A}}
\newcommand{\vZ}{\bm{Z}}

\newcommand{\vz}{\bm{z}}
\newcommand{\vI}{\bm{I}}
\newcommand{\vU}{\bm{U}}
\newcommand{\vD}{\bm{D}}

\newcommand{\MSSA}{\operatorname{\texttt{MSSA}}}

\newcommand{\argmin}{\mathop{\mathrm{arg\,min}}}
\newcommand{\vZero}{\bm{0}}
\newcommand{\softmax}{\operatorname{softmax}}
\newcommand{\ReLU}{\operatorname{ReLU}}

\newcommand{\mat}[1]{\begin{bmatrix}#1\end{bmatrix}}

\usepackage{cleveref}

\title{ Scaling White-Box Transformers for Vision}

\vspace{-.65em}
\author{%
  Jinrui Yang$^{\star 1}$ \,  Xianhang Li$^{\star 1}$  \,  Druv Pai$^2$ \vspace{0.5em}\\
  \textbf{Yuyin Zhou}$^1$ \, \textbf{Yi Ma}$^{2}$ \,  
  \textbf{Yaodong Yu}$^{\dagger 2}$ \,
  \textbf{Cihang Xie}$^{\dagger 1}$ \vspace{.3em}
  \\\small $^{\star}$equal technique contribution,\, $^{\dagger}$equal advising \vspace{.5em} \\
  $^1$UC Santa Cruz \qquad $^2$UC Berkeley
  \vspace{-.65em}
}

\begin{document}

\maketitle

\begin{abstract}
\textsc{crate}, a white-box transformer architecture designed to learn compressed and sparse representations, offers an intriguing alternative to standard vision transformers (ViTs) due to its inherent mathematical interpretability. 
Despite extensive investigations into the scaling behaviors of language and vision transformers, the scalability of \textsc{crate} remains an open question which this paper aims to address. 
Specifically, we propose \textsc{crate}-$\alpha$, featuring strategic yet minimal modifications to the sparse coding block in the \textsc{crate} architecture design, and a light training recipe designed to improve the scalability of \textsc{crate}.
Through extensive experiments, we demonstrate that \textsc{crate}-$\alpha$ can effectively scale with larger model sizes and datasets. 
For example, our \textsc{crate}-$\alpha$-B 
substantially outperforms the prior best \textsc{crate}-B model accuracy on ImageNet classification by 3.7\%, achieving an accuracy of 83.2\%.
Meanwhile, when scaling further, our \textsc{crate}-$\alpha$-L obtains an ImageNet classification accuracy of 85.1\%. 
More notably, these model performance improvements are achieved while preserving, and potentially even enhancing the interpretability of learned \textsc{crate} models, as we demonstrate through showing that the learned token representations of increasingly larger trained \textsc{crate}-$\alpha$ models yield increasingly higher-quality unsupervised object segmentation of images. The project page is \url{https://rayjryang.github.io/CRATE-alpha/}.
\end{abstract}

\section{Introduction}\label{sec:intro}

Over the past several years, the Transformer architecture~\citep{vaswani2017attention} has dominated deep representation learning for natural language processing (NLP), image processing, and visual computing~\citep{devlin2018bert, brown2020language, dosovitskiy2020image, dehghani2023scaling, he2022masked}. 
However, the design of the Transformer architecture and its many variants remains largely empirical and lacks a rigorous mathematical interpretation. 
This has largely hindered the development of new Transformer variants with improved efficiency or interpretability. 
The recent white-box Transformer model \textsc{crate}~\citep{yu2023white} addresses this gap by deriving a simplified Transformer block via unrolled optimization on the so-called \textit{sparse rate reduction} representation learning objective. 

More specifically, layers of the white-box~\textsc{crate} architecture are mathematically derived and fully explainable as unrolled gradient descent-like iterations for optimizing the sparse rate reduction. 
The self-attention blocks of \textsc{crate} explicitly conduct compression via denoising features against learned low-dimensional subspaces, and the MLP block is replaced by an incremental sparsification (via ISTA~\cite{blumensath2008iterative,gregor2010learning}) of the features. 
As shown in previous work~\cite{yu2024emergence}, besides mathematical interpretability, the learned \textsc{crate} models and features also have much better semantic interpretability than conventional transformers, i.e., visualizing features of an image naturally forms a zero-shot image segmentation of that image, even when the model is only trained on classification.

Scaling model size is widely regarded as a pathway to improved performance and emergent properties~\cite{wei2022emergent, touvron2023llama, touvron2023llama2, jiang2023mistral}.
Until now, the deployment of \textsc{crate} has been limited to relatively modest scales. The most extensive model described to date is the base model size encompasses 77.6M parameters (\textsc{crate}-Large)~\cite{yu2023white}. 
This contrasts sharply with standard Vision Transformers (ViTs~\cite{dosovitskiy2020image}), which have been effectively scaled to a much larger model size, namely 22B parameters~\cite{dehghani2023scaling}.

\begin{figure}[t!]
    \centering
   \vspace{-.33em} \includegraphics[width=.99\textwidth]{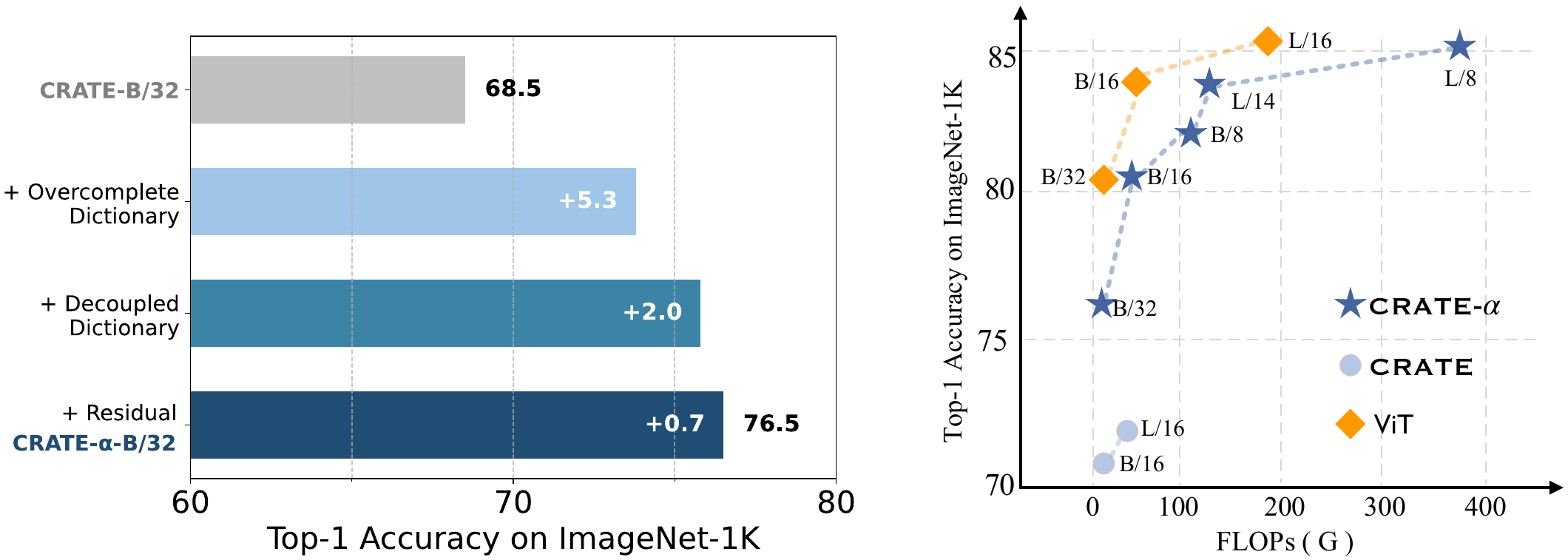}
    \vspace{-0.5em}
    \caption{ \textit{(Left)} We demonstrate how modifications to the components enhance the performance of the \textsc{crate} model. The four models are trained using the same setup: first pre-trained on ImageNet-21K and then fine-tuned on ImageNet-1K. Details
are provided in Section~\ref{sec:crate-alpha}. \textit{(Right)}. 
   We compare the FLOPs and accuracy on ImageNet-1K of our methods with ViT \cite{dosovitskiy2020image} and CRATE \cite{yu2023white}. The values of \ours{} model correspond to those presented in Table~\ref{tab:ours_model_result_on1k}. A more detailed comparison between \ours{} and ViT is included in Appendix \ref{sec:comparison_vit}. }
    \vspace{-1.5em}
\label{fig:ablation_component}
\end{figure}


To this end, this paper provides the first exploration of training \textsc{crate} at different scales for vision, i.e., Tiny, Small, Base, Large, Huge. Detailed model specifications are given in  Table~\ref{tab:model_configs} of Appendix \ref{sec:appendix-model-details}.  To achieve effective scaling, we make two key changes. 
First, we identify the vanilla ISTA block within \textsc{crate} as a limiting factor that hinders further scaling. 
To overcome this, we significantly expand the channels, decouple the association matrix, and add a residual connection, resulting in a new model variant --- \textsc{crate}-$\alpha$. 
It is worth noting that this architecture change still preserves the mathematical interpretability of the model. 
Second, we propose an improved training recipe, inspired by previous work~\cite{touvron2021training,yu2023white,touvron2022deit}, for better coping the training with our new \textsc{crate}-$\alpha$ architecture.

We provide extensive experiments supporting the effective scaling of our \textsc{crate}-$\alpha$ models. 
For example, we scale the \ours{} model from Base to Large size for supervised image classification on ImageNet-21K~\cite{deng2009imagenet}, achieving {\em 85.1\% top-1 accuracy on ImageNet-1K} at the Large model size. 
We further scale the model size from Large to Huge, utilizing vision-language pre-training with contrastive learning on DataComp1B~\citep{gadre2024datacomp}, and achieve {\em a zero-shot top-1 accuracy of 72.3\% on ImageNet-1K} at the Huge model size.\footnote{Model configurations are detailed in \Cref{tab:model_configs} (in \Cref{sec:appendix-model-details}).} 
These results demonstrate the strong scalability of the \ours{} model, shedding light on scaling up mathematically interpretable models for future work.

The main contributions of this paper are threefold:
\vspace{-0.1in}
\begin{enumerate}[leftmargin=*]
\item We design three strategic yet minimal modifications for the \textsc{crate} model architecture to unleash its potential. 
In \Cref{fig:ablation_component}, we reproduce the results of the \textsc{crate} model within our training setup, initially pre-training on ImageNet-21K classification and subsequently fine-tuning on ImageNet-1K classification. 
Compared to the vanilla \textsc{crate} model that achieves 68.5\% top-1 classification accuracy on ImageNet-1K, our \ours{-B/32} model 
significantly improves the vanilla \textsc{crate} model by 8\%, which clearly demonstrates the benefits of the three modifications to the existing \textsc{crate} model. 
Moreover, following the settings of the best \textsc{crate} model and changing the image patch size from 32 to 8, our \ours{-B} model attains a top-1 accuracy of 83.2\% on ImageNet-1K, exceeding the previous best \textsc{crate} model's score of 79.5\% by a significant margin of 3.7\%. 
\vspace{-0.03in}

\item  Through extensive experiments, we show that one can effectively scale \ours{} via model size and data simultaneously. 
In contrast, when increasing the \textsc{crate} model from Base to Large model size, there is a marginal improvement on top-1 classification accuracy (+0.5\%, from 70.8\% to 71.3\%) on ImageNet-1K, indicating diminished returns~\cite{yu2023white}. 
Furthermore, by scaling the training dataset, we achieved a substantial 1.9\% improvement in top-1 classification accuracy on ImageNet-1K, increasing from 83.2\% to 85.1\% when going from \textsc{crate}-$\alpha$ Base to Large.
\vspace{-0.03in}
\item  We further successfully scale \ours{} model from Large to Huge by leveraging vision-language pre-training on DataComp1B. Compared to the Large model, the Huge model (\ours{-H}) achieves a zero-shot top-1 classification accuracy of 72.3\% on ImageNet-1K, marking a significant scaling gain of 2.5\% over the Large model. 
These results indicate that the \textsc{crate} architecture has the potential to serve as an effective backbone for vision-language foundation models.
\end{enumerate}

\subsection*{Related Work}
\textbf{White-box Transformers.} \cite{yu2023white,yu2023white_journal} argued that the quality of a learned representation can be assessed through a unified objective function called the \textit{sparse rate reduction}. 
Based on this framework, \cite{yu2023white,yu2023white_journal} developed a family of transformer-like deep network architectures, named \textsc{crate}, which are mathematically fully interpretable. \textsc{crate} models has been demonstrably effective on various tasks, including vision self-supervised learning and language modeling~\citep{pai2023masked,yu2023white_journal}. 
Nevertheless, it remains unclear whether \textsc{crate} can scale as effectively as widely used black-box transformers. 
Previous work~\cite{yu2023white} suggests that scaling the vanilla \textsc{crate} model can be notably challenging. 

\textbf{Scaling ViT.} 
ViT~\cite{dosovitskiy2020image} represents the initial successful applications of Transformers to the image domain on a large scale.
Many works~\citep{he2022masked,singh2023effectiveness,sun2024eva,sun2023eva,dehghani2023scaling,tolstikhin2021mlp,liu2021swin,liu2022convnet,radford2021learning,li2024inverse,zhai2022scaling} have deeply explored various ways of scaling ViTs in terms of model size and data size. 
From the perspective of self-supervision, MAE~\citep{he2022masked} provides a scalable approach to effectively training a ViT-Huge model using only ImageNet-1K. 
Following the idea of MAE, \cite{singh2023effectiveness} further scales both model parameters to billions and data size to billions of images. 
Additionally, CLIP was the first to successfully scale ViT on a larger data scale (i.e., 400M) using natural language supervision. 
Based on CLIP, \cite{sun2023eva, sun2024eva} further scale the model size to 18 billion parameters, named EVA-CLIP-18B, achieving consistent performance improvements with the scaling of ViT model size. 
From the perspective of supervised learning, \cite{zhai2022scaling, dehghani2023scaling} present a comprehensive analysis of the empirical scaling laws for vision transformers on image classification tasks, sharing some similar conclusions with \cite{kaplan2020scaling}. 
\cite{zhai2022scaling} suggests that the performance-compute frontier for ViT models, given sufficient training data, tends to follow a saturating power law. 
More recently, \cite{dehghani2023scaling} scales up ViT to 22 billion parameters. 
Scaling up different model architectures is non-trivial. 
\cite{tolstikhin2021mlp, liu2021swin, liu2022convnet} have made many efforts to effectively scale up different architectures.  
In this paper, due to the lack of study on the scalability of white-box models, we explore key architectural modifications to effectively scale up white-box transformers in the image domain.

\vspace{-0.1in}
\section{Background and Preliminaries}\label{sec:prelim}
\vspace{-0.1in}
In this section, we present the background on white-box transformers proposed in \citep{yu2023white}, including representation learning objectives, unrolled optimization, and model architecture. 
We first introduce the notation that will be used in the later presentation.

\textbf{Notation.} 
We use notation and problem setup following \citet{yu2023white}. 
We use the matrix-valued random variable $\vX = [\vx_1, \dots, \vx_N] \in \R^{D \times N}$ to represent the data, where each $\vx_i \in \R^D$ is a ``token'', such that each data point is a realization of $\vX$. 
For instance, $\vX$ can represent a collection of image patches for an image, and $\vx_i$ is the $i$-th image patch. 
We use \(f \in \mathcal{F} \colon \R^{D \times N} \to \R^{d \times N}\) to denote the mapping induced by the transformer, and we let \(\vZ = f(\vX) = \mat{\vz_{1}, \dots, \vz_{N}} \in \R^{d \times N}\) denote the features for input data $\vX$. 
Specifically, $\vz_{i}\in\R^{d}$ denotes the feature of the $i$-th input token $\vx_i$. 
The transformer $f$ consists of multiple, say $L$, layers, and so can be written as \(f = f^{L} \circ \cdots \circ f^{1} \circ f^{\textrm{pre}}\), where  \(f^{\ell} \colon \R^{d \times N} \to \R^{d \times N}\) denotes the \(\ell\)-th layer of the transformer, and the pre-processing layer is denoted by \(f^{\textrm{pre}} = \R^{D \times N} \to \R^{d \times N}\). 
The {input} to the \(\ell\)-th layer \(f^{\ell}\) of the transformer is denoted by \(\vZ^{\ell} = \mat{\vz_{1}^{\ell}, \dots, \vz_{N}^{\ell}} \in \R^{d \times N}\), so that \(f^{\ell} \colon \vZ^{\ell} \mapsto \vZ^{\ell + 1}\). In particular, $\vZ^{1} = f^{\textrm{pre}}(\vX) \in \R^{d\times N}$ denotes the output of the pre-processing layer and the input to the first layer.

\subsection{Sparse Rate Reduction}\label{subsec:objective-representation-learning}
\vspace{-0.05in}
Following the framework proposed in \citep{yu2023white_journal}, we posit that the goal of representation learning is to learn a feature mapping or \textit{representation} $f \in \mathcal{F} \colon \R^{D \times N} \to \R^{d \times N}$ that transforms the input data $\vX$ (which may have a nonlinear, multi-modal, and otherwise complicated distribution) into \textit{structured and compact} features $\vZ$, such that the token features lie on a union of low-dimensional subspaces, say with orthonormal bases \(\vU_{[K]} = (\vU_{k})_{k \in [K]} \in (\R^{d \times p})^{K}\). 
\citep{yu2023white} proposes the \textit{\textbf{S}parse \textbf{R}ate \textbf{R}eduction} (SRR) \textit{objective} to measure the goodness of
such a learned representation:
\begin{equation}\label{eq:objective-srr}
     \max_{f\in\mathcal{F}}\, \mathbb{E}_{\vZ = f(\vX)}\left[L_{\texttt{srr}}(\vZ)\right] = \min_{f\in\mathcal{F}}\, \mathbb{E}_{\vZ = f(\vX)}\left[ R^{c}(\vZ\,|\,\vU_{[K]}) - R(\vZ\,|\,\vU_{[K]}) + \lambda \|\vZ\|_{1}  \right],
\end{equation}

\vspace{-0.1in}
where $\vZ = f(\vX)$ denotes the token representation, $\|\vZ\|_1$ denotes the $\ell^{1}$ norm, and $R(\vZ)$, $R^{c}(\vZ\,|\,\vU_{[K]})$ are (estimates for) \textit{rate distortions}~\citep{cover1999elements, derksen2007segmentation}, defined as:
\begin{equation}\label{eq:def-lossy-coding}
    R(\vZ) \doteq \frac{1}{2}\log\det\left(\vI + \frac{d}{N\epsilon^{2}}\vZ^{\top}\vZ\right), \qquad R^{c}(\vZ \mid \vU_{[K]}) \doteq \sum_{k = 1}^{K}R(\vU_{k}^{\top}\vZ).
\end{equation}

In particular, $R^{c}(\vZ \mid \vU_{[K]})$ (resp.~$R(\vZ)$) provide closed-form estimates for the number of bits required to encode the sample $\vZ$ up to precision $\epsilon > 0$, conditioned (resp.~unconditioned) on the samples being drawn from the subspaces with bases $\vU_{[K]}$. Minimizing the term $R^c$ improves the compression of the features $\vZ$ against the posited model, and maximizing the term $R$ promotes non-collapsed features. The remaining term $\lambda \|\vZ\|_{1}$ promotes sparse features. 
Refer to \citep{yu2023white_journal} for more details about the desiderata and objective of representation learning via the rate reduction approach.

\vspace{-0.05in}
\subsection{\textsc{crate}: Coding RATE Transformer}\label{subsec:crate-def}
\vspace{-0.05in}
\textbf{Unrolled optimization.} 
To optimize the learning objective and learn compact and structured representation, one approach is unrolled optimization~\cite{gregor2010learning,tolooshams2021stable}: each layer of the deep network implements an iteration of an optimization algorithm on the learning objective. 
For example, one can design the layer $f^{\ell}$ such that the forward pass is equivalent to a proximal gradient descent step for optimizing learning objective $L(\vZ)$, i.e., $\vZ^{\ell+1} = f^{\ell}(\vZ^{\ell}) = \texttt{Prox}[\vZ^{\ell} - \eta\cdot\nabla_{\vZ} L(\vZ^{\ell})]$. Here we use $\eta$ to denote the step size and $\texttt{Prox}[\cdot]$ to denote the proximal operator~\cite{parikh2014proximal}.

\textbf{One layer of the \textsc{crate} model.} We now present the design of each layer of the white-box transformer architecture -- Coding RATE Transformer (\textsc{crate}) -- proposed in \cite{yu2023white}.
Each layer of \textsc{crate} contains two blocks: the compression block and the sparsification block. These correspond to a two-step alternating optimization procedure for optimizing the sparse rate reduction objective \eqref{eq:objective-srr}. Specifically, the $\ell$-th layer of \textsc{crate} is defined as
\begin{equation}
    \vZ^{\ell+1} = f^{\ell}(\vZ^{\ell}) = {\texttt{ISTA}(\vZ^{\ell+1/2}\,|\,\vD^{\ell})}, \quad \mathrm{where}\quad \vZ^{\ell + 1/2}
    = \vZ^{\ell} + \MSSA(\vZ^{\ell}).
\end{equation}
\textbf{{Compression block (MSSA).}} 
The compression block in \textsc{crate}, called \textbf{M}ulti-head \textbf{S}ubspace \textbf{S}elf-\textbf{A}ttention block (\(\MSSA\)), is derived for compressing the token set $\vZ=[\vz_1, \dots, \vz_{N}]$ by optimizing the compression term $R^c$ (defined Eq.~\eqref{eq:objective-srr}), i.e.,
\begin{align}\label{eq:grad_rc_mssa}
    \vZ^{\ell + 1/2}
    &= \vZ^{\ell} + \MSSA(\vZ^{\ell} \mid \vU_{[K]}^{\ell}) \approx \vZ^{\ell} - \kappa \nabla_{\vZ}R^{c}(\vZ^{\ell}\,|\,\vU_{[K]}^{\ell}),
\end{align}
where $\vU_{[K]}^{\ell}$ denotes the (local) signal model at layer $\ell$, and the \(\MSSA\) operator is defined as
\begin{equation}\label{eq:mssa}
    \MSSA(\vZ \mid \vU_{[K]}) =  \frac{\kappa p}{N\epsilon^{2}}\,[\vU_{1} \,\cdots\, \vU_{K}]\mat{(\vU_{1}^{\top}\vZ)\softmax((\vU_{1}^{\top}\vZ)^{\top}(\vU_{1}^{\top}\vZ)) \\ \vdots \\ (\vU_{K}^{\top}\vZ)\softmax((\vU_{K}^{\top}\vZ)^{\top}(\vU_{K}^{\top}\vZ))}.
\end{equation}
Compared with the commonly used attention block in transformer~\cite{vaswani2017attention}, where the $k$-th attention head is defined as $(\bm{V}_{k}^{\top}\vZ)\softmax((\bm{Q}_{k}^{\top}\vZ)^{\top}(\bm{K}_{k}^{\top}\vZ))$, $\MSSA$ uses only one matrix to obtain the query, key, and value matrices in the attention: that is,  $\vU_{k} = \bm{Q}_{k} = \bm{K}_{k} = \bm{V}_{k}$.

\textbf{{Sparse coding block (ISTA).}} 
The Iterative Shrinkage-Thresholding Algorithm (ISTA) block is designed to optimize the sparsity term and the global coding rate term, $ \lambda \|\vZ\|_{0} - R(\vZ \mid \vU_{[K]})$ in \eqref{eq:objective-srr}. 
\cite{yu2023white} shows that an optimization strategy for these terms posits a (complete) incoherent dictionary $\vD^{\ell} \in \R^{d \times d}$ and takes a proximal gradient descent step towards solving the associated LASSO problem $\argmin_{\vZ \geq \vZero}[\frac{1}{2}\|{\vZ^{\ell + 1/2} - \vD^{\ell}\vZ}\|_{2}^{2} + \lambda \|{\vZ}\|_{1}]$, obtaining the iteration
\begin{align}\label{eq:grad_lasso_ista}
    \vZ^{\ell + 1}
    &= \texttt{ISTA}(\vZ^{\ell+1/2}\,|\,\vD^{\ell}) = \mathrm{ReLU}(\vZ^{\ell+1/2} + \eta\, (\vD^{\ell})^{\top}(\vZ^{\ell+1/2} - \vD^{\ell}\vZ^{\ell+1/2}) - \eta\lambda\mathbf{1}).
\end{align}
In particular, the \texttt{ISTA} block sparsifies the intermediate iterates $\vZ^{\ell + 1/2}$ w.r.t.~$\vD^{\ell}$ to obtain $\vZ^{\ell + 1}$.

\begin{figure}[t]
    \centering
    \includegraphics[width=.85\textwidth]{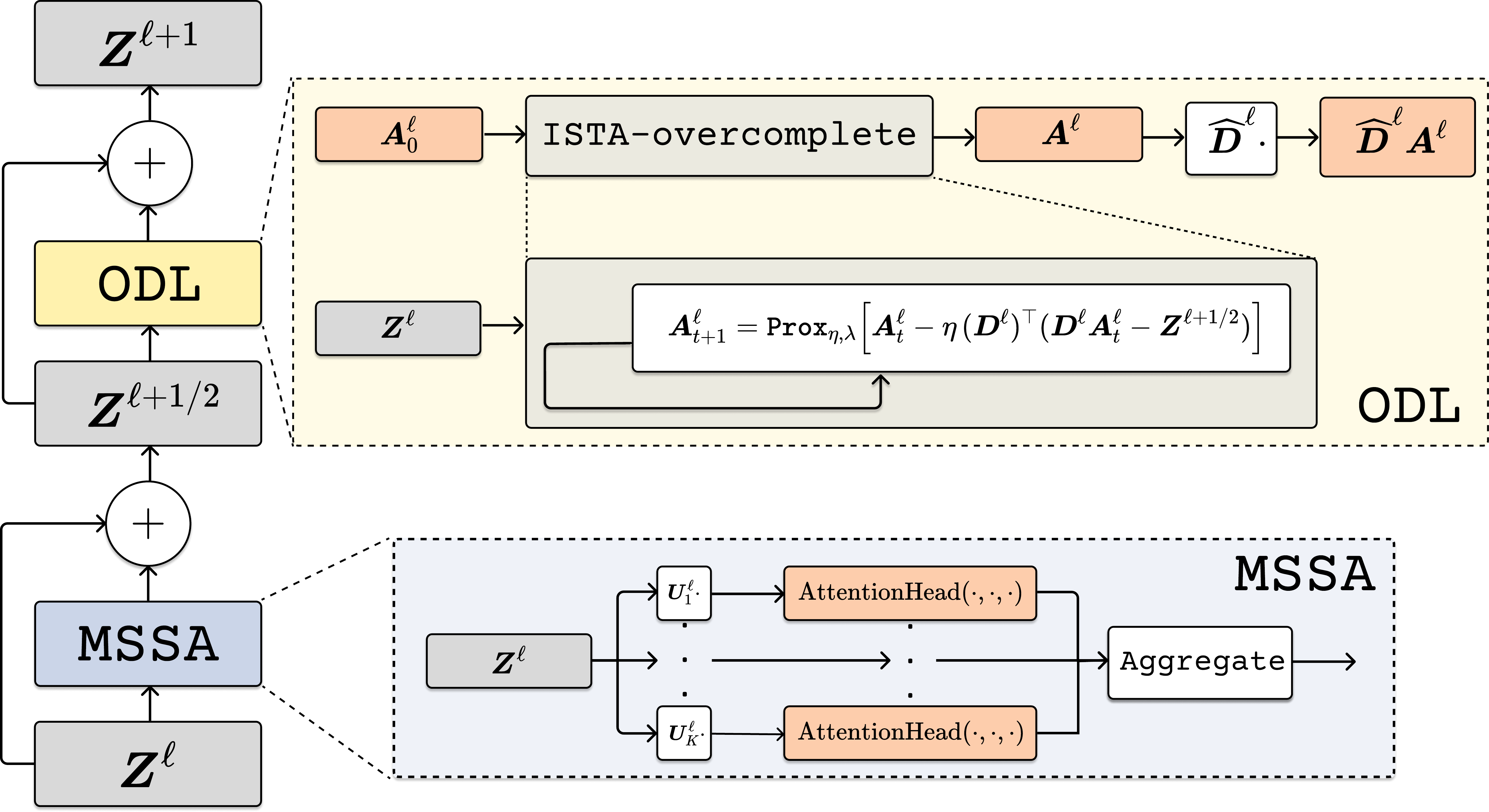}
    \caption{ One layer of the \ours{} model architecture. $\MSSA$ (\textbf{M}ulti-head \textbf{S}ubspace \textbf{S}elf-\textbf{A}ttention, defined in \eqref{eq:mssa}) represents the compression block, and \texttt{ODL} (\textbf{O}vercomplete \textbf{D}ictionary \textbf{L}earning, defined in \eqref{eq:odl}) represents the sparse coding block. A more detailed illustration of the modifications is provided in Fig.~\ref{fig:exp-rc-sparisty-small-new} in the Appendix . }
    \vspace{-1.5em}
    \label{fig:framework}
\end{figure}

\section{CRATE-$\alpha$ Model}\label{sec:crate-alpha}
In this section, we present the \textsc{crate}-$\alpha$ architecture, which is a variant of \textsc{crate}~\cite{yu2023white}. 
As shown in Fig.~\ref{fig:ablation_component} (\textit{Right}), there is a significant performance gap between the white-box transformer \textsc{crate}-B/16 (70.8\%) and the vision transformer ViT-B/16 (84.0\%)~\cite{dosovitskiy2020image}. 
One possible reason is that the $\texttt{ISTA}$ block applies a complete dictionary $\vD \in \R^{d\times d}$, which may limit its expressiveness. In contrast, the MLP block in the transformer\footnote{The MLP block is defined as $\vZ^{\ell+1} = \vZ^{\ell} + \bm{W}_{2}\sigma(\bm{W}_{1}^{\top}\vZ^{\ell+1/2})$, where $\sigma$ is the nonlinear activation function and $\vZ^{\ell+1/2}$ denotes the output of the attention block.} applies two linear transformations $\bm{W}_{1}, \bm{W}_{2} \in \R^{d \times 4d}$, leading to the MLP block having 8 times more parameters than the $\texttt{ISTA}$ block.

Since the ISTA block in CRATE applies a single incremental step to optimize the sparsity objective, applying an orthogonal dictionary can make it ineffective in sparsifying the token representations.
Previous work~\cite{qu2019geometric} has theoretically demonstrated that overcomplete dictionary learning enjoys a favorable optimization landscape.
In this work, we use an overcomplete dictionary in the sparse coding block to promote sparsity in the features. 
Specifically, instead of using a complete dictionary $\vD^{\ell}\in\R^{d\times d}$, we use an overcomplete dictionary $\vD^{\ell}\in \R^{d \times (Cd)}$, where $C>1$ (a positive integer) is the overcompleteness parameter. 
Furthermore, we explore two additional modifications to the sparse coding block that lead to improved performance for \textsc{crate}. 
We now describe the three variants of the sparse coding block that we use in this paper.

\textbf{Modification~\#1: Overparameterized sparse coding block.} 
For the output of the $\ell$-th \textsc{crate} attention block $\vZ^{\ell+1/2}$, we propose to sparsify the token representations with respect to an overcomplete dictionary $\vD^{\ell}\in\R^{d\times (Cd)}$ by optimizing the following LASSO problem,
\begin{equation}\label{eq:lasso-overcomplete}
    \vA^{\ell}
    \approx \argmin_{\vA \geq \vZero}\Big[\frac{1}{2}\|{\vZ^{\ell + 1/2} - \vD^{\ell}\vA}\|_{2}^{2} + \lambda \|{\vA}\|_{1}\Big].
\end{equation}
To approximately solve \eqref{eq:lasso-overcomplete}, we apply two steps of proximal gradient descent, i.e.,
\begin{equation}\label{eq:grad_ista_1}
    \vA_{0}^{\ell} = \vZero, \qquad \vA^{\ell}_{1} = \texttt{Prox}_{\eta,\lambda}\big[\vA^{\ell}_{0}; \vD^{\ell}, \vZ^{\ell+1/2}\big], \qquad
    \vA^{\ell}_{2}
    = \texttt{Prox}_{\eta,\lambda}\big[\vA^{\ell}_{1}; \vD^{\ell}, \vZ^{\ell+1/2}\big],
\end{equation}
where $\texttt{Prox}$ is the proximal operator of the above non-negative LASSO problem \eqref{eq:lasso-overcomplete} and defined as
\begin{equation}
    \texttt{Prox}_{\eta, \lambda}[\vA; \vD, \vZ] = \ReLU(\vA - \eta \vD^{\top}(\vD\vA - \vZ) - \eta \lambda \bm{1}).
\end{equation}
The output of the sparse coding block is defined as
\begin{equation}\label{eq:sparse_code_def}
    \vZ^{\ell + 1}
    = \vD^{\ell}\vA^{\ell}, \quad \mathrm{where}\quad \vA^{\ell} = \vA^{\ell}_2 \doteq \texttt{ISTA-OC}(\vZ^{\ell + 1/2} \mid \vD^{\ell}).
\end{equation}
Namely, $\vA^{\ell}$ is a sparse representation of $\vZ^{\ell+1/2}$ with respect to the overcomplete dictionary $\vD^{\ell}$.
The original \textsc{crate} \texttt{ISTA} tries to learn a complete dictionary $\vD\in\R^{d\times d}$ to transform and sparsify the features $\vZ$. 
By leveraging more atoms than the ambient dimension, the overcomplete dictionary $\vD\in\R^{d\times (Cd)}$ can provide a redundant yet expressive codebook to identify the salient sparse structures underlying $\vZ$. 
As shown in Fig.~\ref{fig:ablation_component}, the overcomplete dictionary design leads to $5.3\%$ improvement compared to the vanilla \textsc{crate} model.

\textbf{Modification~\#2: Decoupled dictionary.} 
We propose to apply a decoupled dictionary $\widehat{\vD}^{\ell}$ in the last step (defined in \eqref{eq:sparse_code_def} of the sparse coding block, $\vZ^{\ell + 1}
= \widehat{\vD}^{\ell} \vA^{\ell}$,
where $\widehat{\vD}^{\ell}  \in \R^{d \times (Cd)}$ is a different dictionary compared to $\vD^{\ell}$. 
By introducing the decoupled dictionary, we further improve the model performance by $2.0\%$, as shown in Fig.~\ref{fig:ablation_component}. We denote this mapping from \(\vZ^{\ell + 1/2}\) to \(\vZ^{\ell + 1}\) as the \textbf{O}vercomplete \textbf{D}ictionary \textbf{L}earning block (\texttt{ODL}), defined as follows: 
\begin{equation}
    \texttt{ODL}(\vZ^{\ell + 1/2} \mid \vD^{\ell}, \widehat{\vD}^{\ell}) \doteq \widehat{\vD}^{\ell}\cdot\texttt{ISTA-OC}(\vZ^{\ell + 1/2} \mid \vD^{\ell}) = \widehat{\vD}^{\ell}\vA^{\ell}.
\end{equation}

\textbf{Modification~\#3: Residual connection.} Based on the previous two modifications, we further add a residual connection, obtaining the following modified sparse coding block:
\begin{equation}\label{eq:odl}
    \vZ^{\ell + 1} = \vZ^{\ell + 1/2} +\texttt{ODL}(\vZ^{\ell + 1/2} \mid \vD^{\ell}, \widehat{\vD}^{\ell}).
\end{equation}

An intuitive interpretation of this modified sparse coding block is as follows: instead of directly sparsifying the feature representations $\vZ$, we first identify the potential sparse patterns present in $\vZ$ by encoding it over a learned dictionary. Subsequently, we incrementally refine $\vZ$ by exploiting the sparse codes obtained from the previous encoding step. 
From Fig.~\ref{fig:ablation_component}, we find that the residual connection leads to a $0.7\%$ improvement.

To summarize, to effectively scale white-box transformers, we implement three modifications to the vanilla white-box \textsc{crate} model proposed in \cite{yu2023white}. Specifically, in our \textsc{crate}-$\alpha$ model, we introduce a decoupling mechanism, quadruple the dimension of the dictionary (4$\times$), and incorporate a residual connection in the sparse coding block.

\begin{figure}[t]
    \centering
    \includegraphics[width=.495\textwidth]{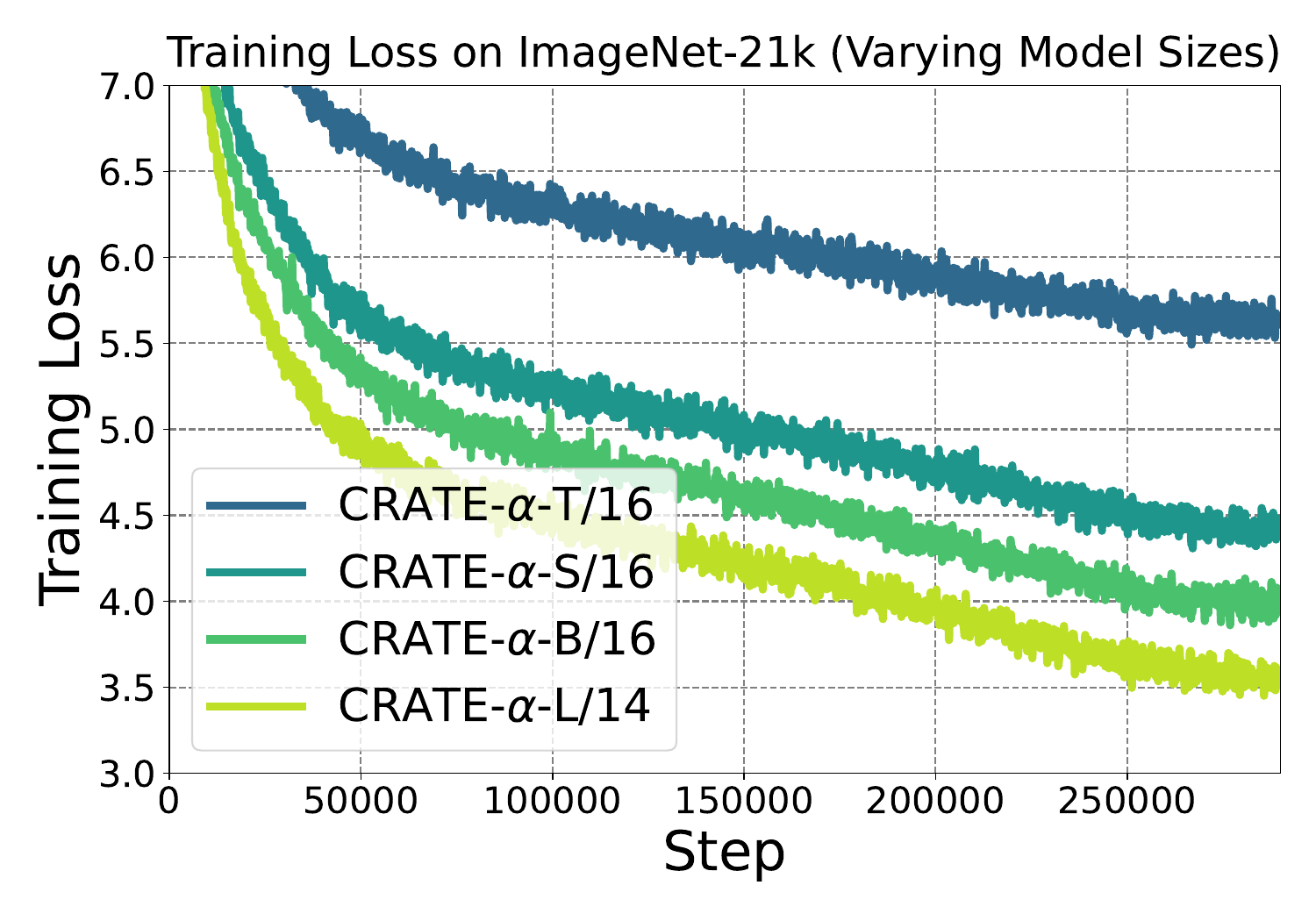}
    \includegraphics[width=.495\textwidth]{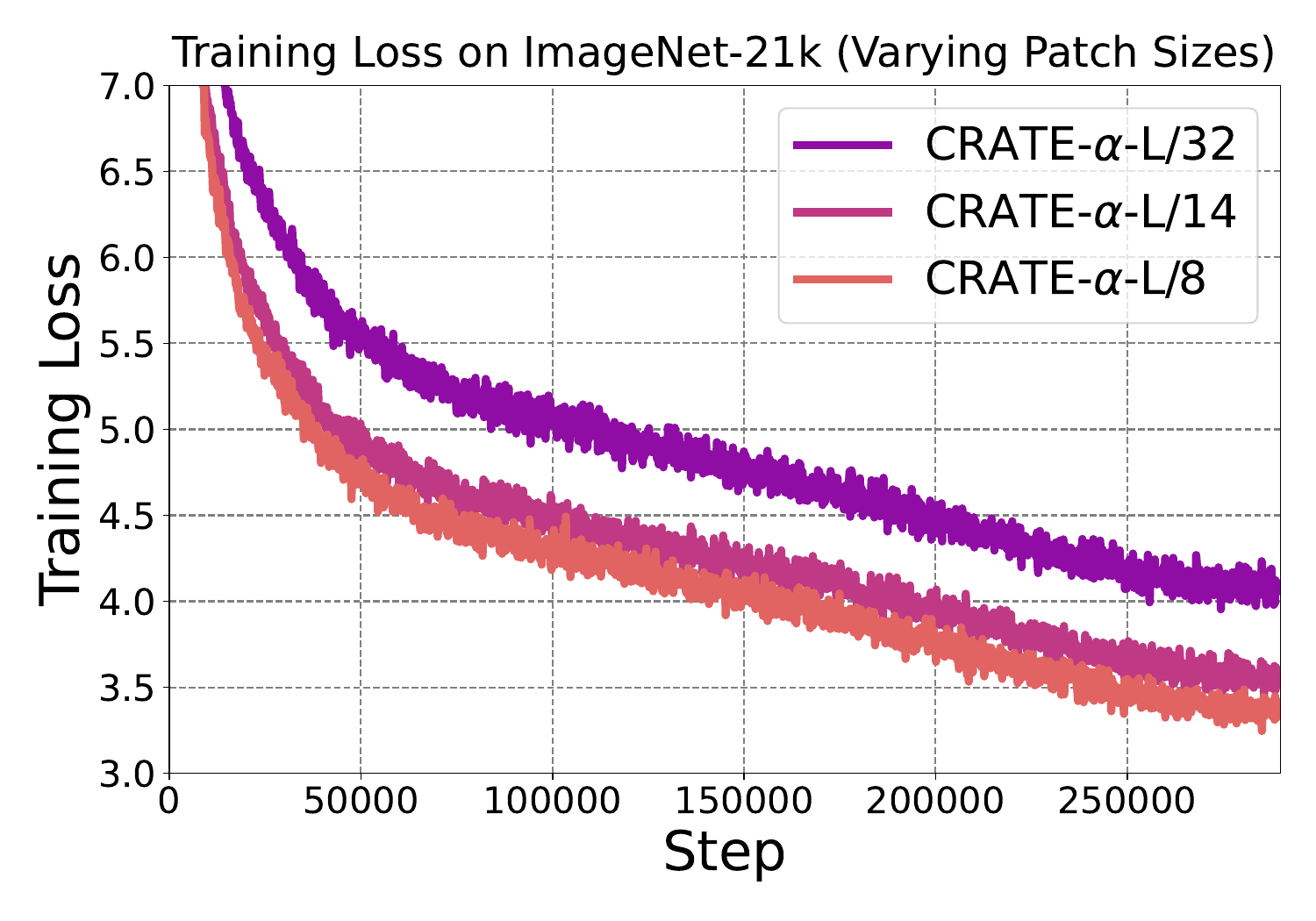}
    \caption{Training loss curves of \ours{} on ImageNet-21K.
    (\textit{Left}) Comparing training loss curves across \ours{} with different model sizes. 
    (\textit{Right}) Comparing training loss curves across \ours{-Large} with different patch sizes.}
    \vspace{-1em}
    \label{fig:training_loss_curves}
\end{figure}

\vspace{-0.05in}
\section{Experiments}
\label{sec:Experiments}
\vspace{-0.05in}

\textbf{Overall.} The experimental section consists of three parts: (1) \textbf{Scaling  study:} We thoroughly investigate the scaling behaviors of \ours{} from Base to Large size and ultimately to Huge size. (2) \textbf{Downstream applications:} To further verify the broader benefits of scaling the \ours{} model, we conduct additional experiments on real-world downstream tasks and present preliminary exploration results of \ours{} on language tasks. (3) \textbf{Interpretability:} In addition to scalability, we study the interpretability of \ours{} across different model sizes.

\subsection{Dataset and Evaluation}

\textbf{Scaling Study.} For the transition from Base to Large size, we pre-train our model on ImageNet-21K and fine-tune it on ImageNet-1K via supervised learning. When scaling from Large to Huge, we utilize the DataComp1B~\cite{gadre2024datacomp} dataset within a vision-language pre-training paradigm, allowing us to study the effects of scaling the model to a massive size. For evaluation, we evaluate the zero-shot accuracy of these models on ImageNet-1K.

\textbf{Downstream Applications.} We include additional experimental results on four downstream datasets (CIFAR-10/100, Oxford Flowers, and Oxford-IIT Pets). We also examine the dense prediction capability of \ours{} by training it on segmentation tasks using the ADE20K dataset~\cite{zhou2017scene}. For language tasks, we conduct new experiments with \ours{} using autoregressive training on OpenWebText, following the setup in nanoGPT~\cite{author_github_repo}.

\textbf{Interpretability}. Following the evaluation setup of \textsc{crate} as outlined in \cite{yu2023white}, we apply MaskCut~\cite{wang2023cut} to validate and evaluate the rich semantic information captured by our model in a zero-shot setting, including both qualitative and quantitative results.

\subsection{Training and Fine-tuning Procedures}

\textbf{Scaling Study.} 
\textbf{(1) Base to Large size:}  We initially pre-train the  \textsc{crate}-$\alpha$ model on ImageNet-21K and subsequently fine-tune it on ImageNet-1K.
During the pre-training phase, we set the learning rate to $8 \times 10^{-4}$,
weight decay to 0.1, and batch size to 4096. 
We apply data augmentation techniques such as Inception crop~\cite{szegedy2015going} resized to 224 and random horizontal flipping. 
In the fine-tuning phase, we adjust the base learning rate to $1.6\times 10^{-4}$, maintain weight decay at 0.1, and batch size at 4096. 
We apply label smoothing with a smoothing parameter of 0.1 and apply data augmentation methods including Inception crop, random horizontal flipping, and random augmentation with two transformations (magnitude of 9). 
For evaluation, we resize the smaller side of an image to 256 while maintaining the original aspect ratio and then crop the central portion to 224$\times$224. 
In both the pre-training and fine-tuning phases, we use the AdamW optimizer~\cite{loshchilov2017decoupled} and incorporate a warm-up strategy, characterized by a linear increase over 10 epochs. 
Both the pre-training and fine-tuning are conducted for a total of 91 epochs, utilizing a cosine decay schedule.  

\textbf{(2) Large to Huge size:}  In the pre-training stage, we utilize an image size of 84$\times$84, and the maximum token length is 32, with a total of 2.56 billion training samples. 
During the fine-tuning stage, we increase the image size to 224$\times$224 while maintaining the maximum token length at 32, with a 512 million training samples. 
Here, the key distinction between the pre-training stage and the fine-tuning stage is the image size. A smaller image size results in a faster training speed. 
In the configurations of \ours{}-CLIPA-B, \ours{}-CLIPA-L, and \ours{}-CLIPA-H, we use the \ours{} model as the vision encoder, and utilize the same pre-trained huge transformer model from CLIPA~\cite{li2024inverse} as the text encoder. 
For both the pre-training and fine-tuning stages, we freeze the text encoder and only train the vision encoder, i.e., the \ours{} model. 
As we will show in the later results, this setup effectively demonstrates the scaling behaviors of \ours{} models in the image domain. 
{ Detailed hyperparameter settings can be found in \Cref{sec:appendix-details}}.

\textbf{Downstream Applications.} On four downstream datasets, we follow the training setup from~\cite{yu2023white}. For the segmentation task, we compare the performance of CRATE and \ours{} on the ADE20K dataset, mainly following the setup of~\cite{ren2023tinymim} with minor modifications. Our batch size is set to 128, and the total number of training steps is 5000.  For the language task, we conduct experiments with \ours{} using autoregressive training on OpenWebText, following the setup in \cite{author_github_repo}. We compare \ours{} models (57M and 120M) with CRATE and GPT-2, using results from CRATE reported in~\cite{yu2023white_journal}.

\begin{table*}[t!]
\caption{Top-1 accuracy of \ours{} on ImageNet-1K with different model scales when pre-trained on ImageNet-21K and then fine-tuned on ImageNet-1K. For comparison, we also list the results from the paper \cite{yu2023white} which demonstrate the diminished return from  \textsc{crate} base to large, trained only on ImageNet-1K. "IN-21K" refers to ImageNet-21K. ($^{\ddagger}$Results from \cite{yu2023white}.)
 }
\vspace{-0.05in}
\centering
\small
    \setlength{\tabcolsep}{12pt}

\begin{tabular}{@{}lc|ccc@{}}
\toprule
 Models (Base) & IN-1K(\%)  & Models (Large) & IN-1K(\%)  \\ 
\midrule
\midrule
  {\color{gray}  \textsc{crate}-B/16 w/o IN-21K}  & {\color{gray} 70.8$^{\ddagger}$ } & {\color{gray}  \textsc{crate}-L/16 w/o IN-21K}  &  {\color{gray}71.3$^{\ddagger}$} \\
 \midrule
\midrule
 \ours{-B/32}  & 76.5 & \ours{-L/32} &  80.2 \\
\midrule
 \ours{-B/16}  & 81.2 & \ours{-L/14}
 &  83.9 \\
 \midrule
 \ours{-B/8}  & 83.2 & \ours{-L/8} &  85.1 \\
 \bottomrule
\end{tabular}%
\label{tab:ours_model_result_on1k}
\vspace{-0.15in}
\end{table*}

\begin{figure}[t!]
    \centering
        \begin{subfigure}[ht]{0.495\textwidth}
        \centering
        \includegraphics[width=\textwidth]{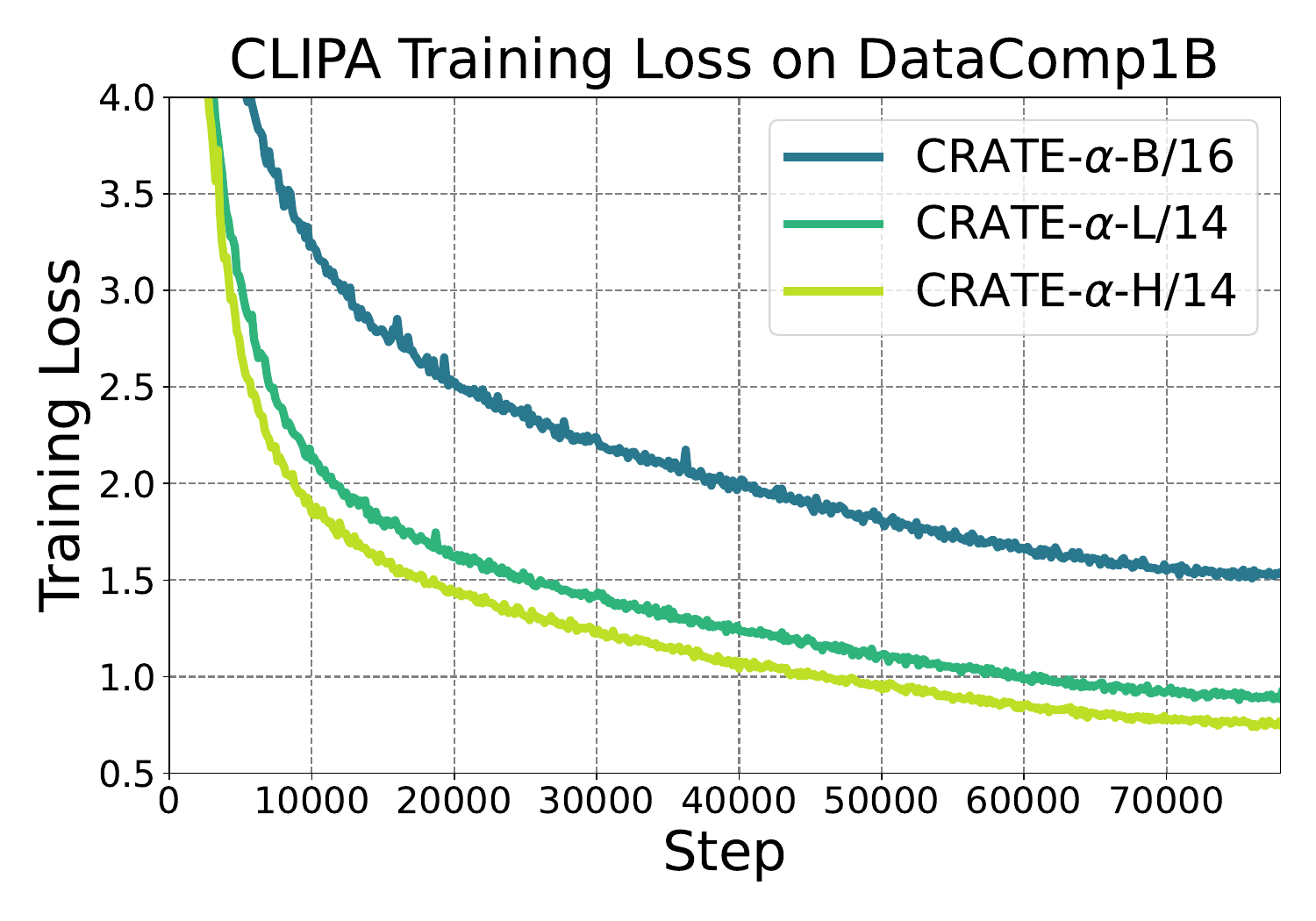}
    \end{subfigure}
    \hfill
    \begin{subfigure}[ht]{0.495\textwidth}
        \centering
        \includegraphics[width=\textwidth]{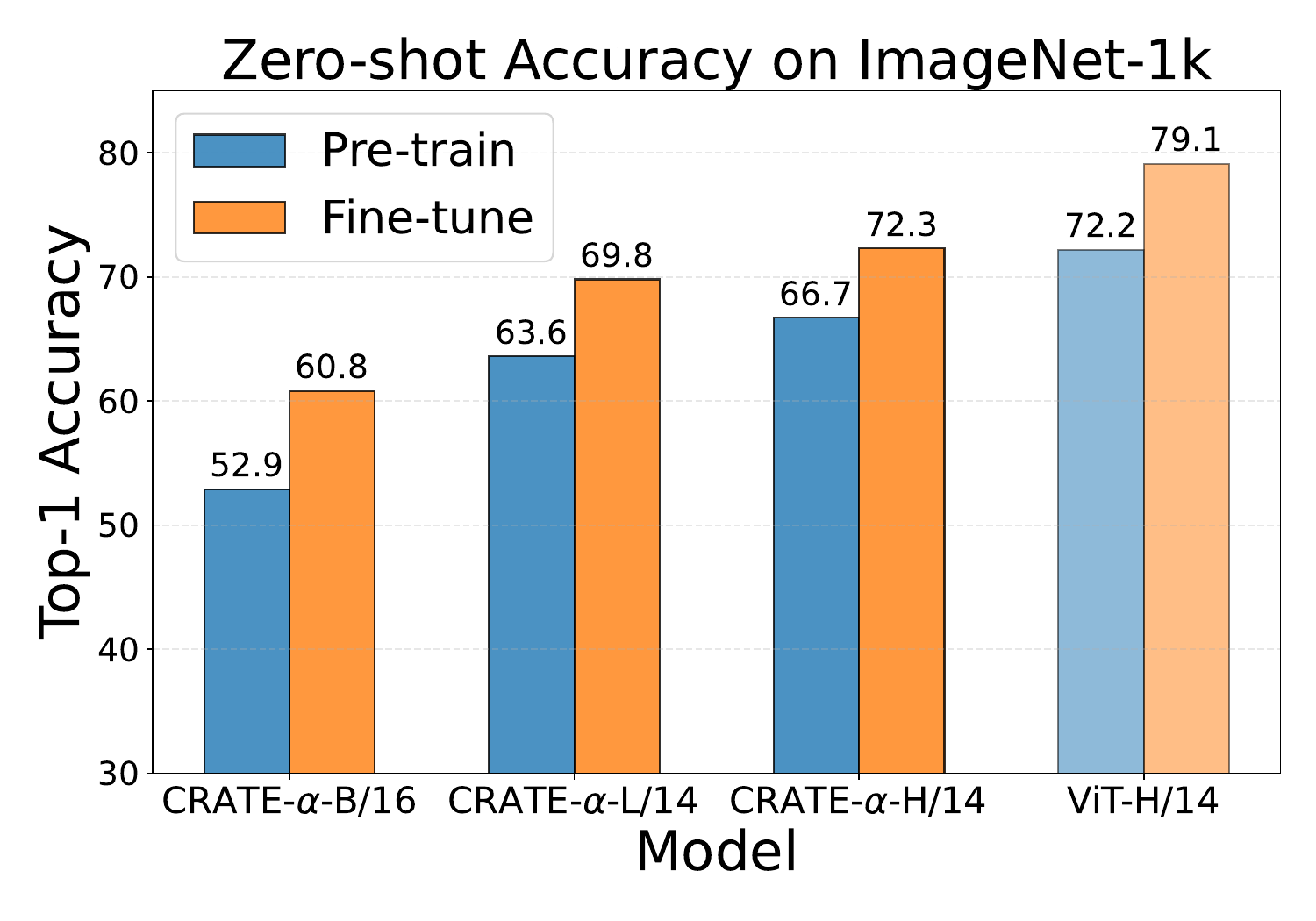}
    \end{subfigure}
    \vspace{-1em}
    \caption{(\textit{Left}) Comparing training loss curves of \ours{-CLIPA} with different model sizes on DataComp1B. 
    (\textit{Right}) Comparing zero-shot accuracy of \ours{-B/L/H} models and ViT-H on ImageNet-1K.
    }
    \vspace{-0.5em}
    \label{fig:clipa_results}
\end{figure}

\subsection{Results and Analysis}

\textbf{Scaling the \ours{} Model from Base to Large.} As shown in Table~\ref{tab:ours_model_result_on1k}, we compare \ours{-B} and \ours{-L} at patch sizes 32, 16, and 8. 
Firstly, we find our proposed \ours{-L} consistently achieves significant improvements across all patch sizes. 
Secondly, compared with the results of the vanilla  \textsc{crate} (the first row of Table~\ref{tab:ours_model_result_on1k}), increasing from  \textsc{crate}-B to  \textsc{crate}-L results in only a 0.5\% improvement on ImageNet-1K. This indicates a case of diminishing returns.
These findings compellingly highlight that the scalability of \ours{} models significantly outperforms that of the vanilla  \textsc{crate}. Meanwhile, the training loss in the pre-training stage is presented in Fig.~\ref{fig:training_loss_curves}; as the model capacity increases, the trend of the training loss improves predictably. This phenomenon is also described in \cite{dosovitskiy2020image}.

\textbf{Scaling the \ours{} Model from Large to Huge.}  From the results shown in Fig.~\ref{fig:clipa_results}, we find that: (1)\,\ours{}-CLIPA-L/14 significantly outperforms \ours{}-CLIPA-B/16 by 11.3\% and 9.0\% in terms of ImageNet-1K zero-shot accuracy during the pre-training and fine-tuning stages, respectively. 
The substantial benefit suggests that the quality of learned representation may be constrained by the model size. 
Therefore, increasing the model size effectively leverages larger amounts of data.
(2) When continuing to scale up model size, we also observe that \ours{}-CLIP-H/14 continues to benefit from larger training datasets, outperforming \ours{}-CLIP-L/14 by 3.1\% and 2.5\% in terms of ImageNet-1K zero-shot accuracy during the pre-training and fine-tuning stages, respectively. 
This demonstrates the strong scalability of the \ours{} model. 
To explore the performance ceiling, we train a standard ViT-CLIPA-H/14 from scratch and observe improved performance.

\textbf{Downstream Applications.} On four downstream datasets, as shown in Table \ref{tab:dataset_comparison}, we find that \ours{} consistently outperforms CRATE, with both models pre-trained on IN21K, while \ours{} demonstrates improved performance as model size increases. For the segmentation task, results in Table \ref{tab:segmentation} show that \ours{} consistently outperforms CRATE across all key metrics, with both models pre-trained on IN21K. These findings indicate significant performance gains in vision tasks beyond classification. For the language task, Table \ref{tab:cross_entropy_loss_comparison} shows that \ours{} significantly improves over CRATE in language modeling. Due to limited time and resource constraints, we completed 80\% of the total iterations for \ours{}-small and 55\% for \ours{}-base, compared to the 600K total iterations used for CRATE. Nevertheless, \ours{} still demonstrated notable improvements.

\begin{figure}[t]
    \centering
    \includegraphics[width=1.0\textwidth]{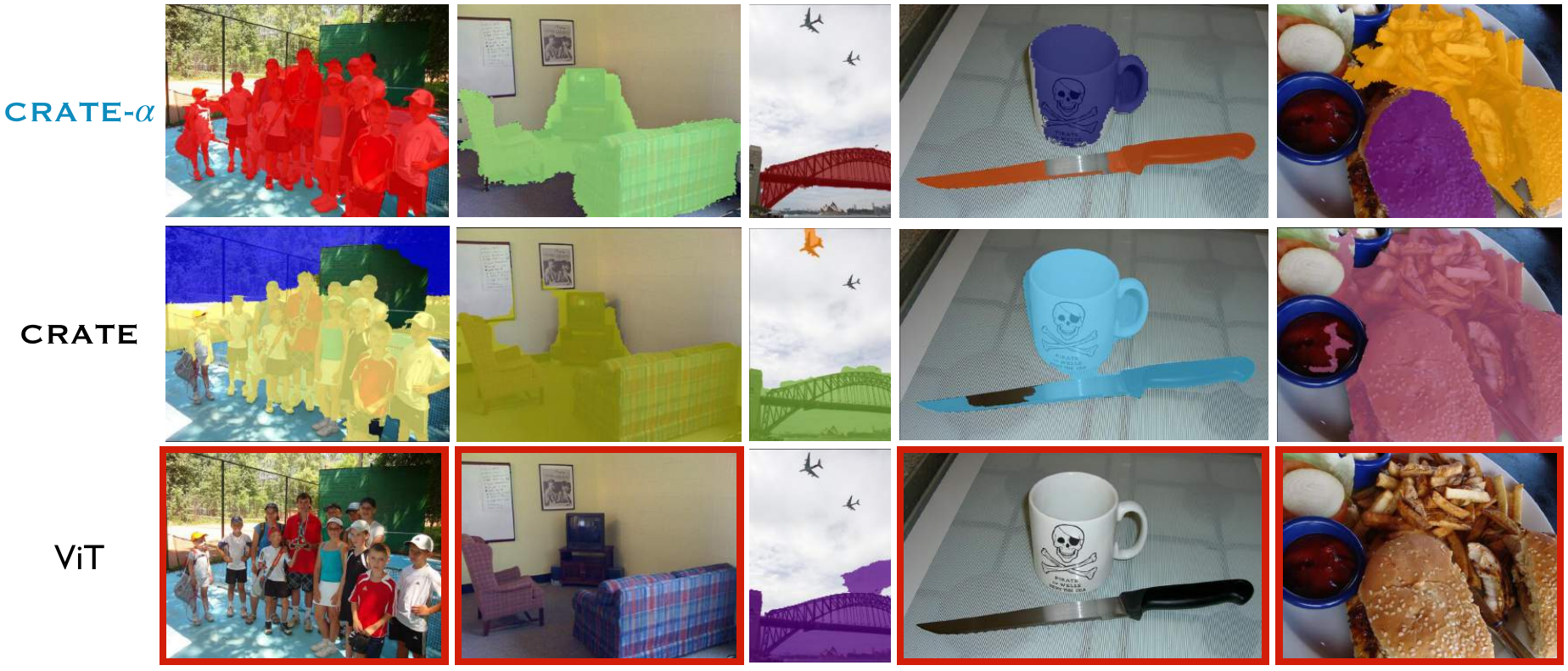}
  \caption{\textbf{Visualization of segmentation on COCO {val2017}~\cite{lin2014microsoft} with MaskCut~\cite{wang2023cut}.} 
    (\textit{Top row}) Supervised \ours{} effectively identifies the main objects in the image. Compared with  \textsc{crate} (\textit{Middle row}), \ours{} achieves better segmentation performance in terms of boundary.
    (\textit{Bottom row}) Supervised ViT fails to identify the main objects in most images. We mark failed image with \includegraphics[width=.25cm]{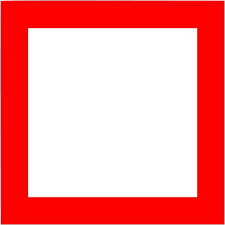}\,.}
    \label{fig:cutler_vis}
\end{figure}

\textbf{Interpretability.} As shown in Fig.~\ref{fig:cutler_vis}, we provide the segmentation visualization on COCO val2017~\cite{lin2014microsoft} for \ours{}, \textsc{crate}, and ViT, respectively. 
We find that our model preserves and even improves the (semantic) interpretability advantages of \textsc{crate}. 
Moreover, we summarize quantitative evaluation results on COCO val2017 in Table~\ref{tab: maskcut_main_result}. 
Interestingly, when scaling up model size for \ours{}, the Large model improves over the Base model in terms of object detection and segmentation.

\begin{table*}[htb!]
\caption{The performance comparison between CRATE and CRATE-$\alpha$ across various datasets.}
\vspace{-0.05in}
\centering
\small
\setlength{\tabcolsep}{2pt} 
\resizebox{0.95\textwidth}{!}{
\begin{tabular}{lccccc}
\toprule
Dataset & CRATE-B/32 & CRATE-$\alpha$-B/32 & CRATE-$\alpha$-L/32 & CRATE-$\alpha$-B/16 & CRATE-$\alpha$-L/14 \\ 
\midrule
CIFAR-10           & 97.22 & 98.17 & 98.68 & 98.67 & 99.10 \\
CIFAR-100          & 85.27 & 89.40 & 91.16 & 90.58 & 92.57 \\
Oxford Flowers-102 & 93.90 & 97.77 & 99.01 & 99.27 & 99.56 \\
Oxford-IIIT-Pets   & 80.38 & 88.19 & 90.46 & 92.70 & 93.98 \\
\bottomrule
\end{tabular}%
}
\label{tab:dataset_comparison}
\vspace{-0.15in}
\end{table*}

\begin{table}[h]
\caption{Performance comparison of CRATE models with different configurations.}
\centering
\begin{tabular}{lcccc}
\toprule
Model & Scope & mIoU & mAcc & aAcc \\
\midrule
\ours{}-B/32 & global & 35.35 & 45.28 & 77.63 \\
CRATE-B/32   & global & 30.28 & 39.29 & 75.21 \\
\bottomrule
\end{tabular}
\label{tab:segmentation}
\end{table}

\begin{table}[htb!]
\caption{The comparison between CRATE and CRATE-$\alpha$ on the NLP task using the OpenWebText dataset.}
\centering
\small
\setlength{\tabcolsep}{6pt} 
\resizebox{0.95\textwidth}{!}{
\begin{tabular}{lcccc}
\toprule
 & GPT-2-base & CRATE-base & CRATE-$\alpha$-small & CRATE-$\alpha$-base \\ 
\midrule
Model size & 124M & 60M & 57M & 120M \\
CE val loss & 2.85 & 3.37 & 3.28 & 3.14 \\
\bottomrule
\end{tabular}%
}
\label{tab:cross_entropy_loss_comparison}
\vspace{-0.15in}
\end{table}

\vspace{-0.05in}
\subsection{Compute-efficient Scaling Strategy }
\label{sec:efficient_scaling}
\vspace{-0.05in}
We further explore methods to scale models efficiently in terms of computation. 
Table~\ref{tab:ours_model_result_on1k} demonstrates that the \ours{} model scales effectively from the Base model to its larger variants. 
However, the pre-training computation for the top-performing model, \ours{-L/8}, is resource-intensive on ImageNet-21K.
Inspired by CLIPA~\cite{li2024inverse}, we aim to reduce computational demands by using reduced image token sequence lengths, while maintaining the same training setup during the fine-tuning stage. The results are summarized in  Table~\ref{tab:compute_efficient_scaling}.

\begin{table*}[t!]
\caption{ Compute-efficient scaling strategy. To reduce the compute requirements of the pre-training stage, we use a model with a larger patch size. This results in a shorter token length for the same input size. The second and fourth columns indicate the compute requirements for the pre-training and fine-tuning stages, respectively, measured in TPU v3 core-hours. 
Details are provided in Section~\ref{sec:efficient_scaling}.}
\centering
\small
    \setlength{\tabcolsep}{10pt}

\begin{tabular}{@{}lc|lc|ccc@{}}
\toprule
   Pre-train & Core-hours   &  Fine-tune & Core-hours & Total core-hours & IN-1K(\%)  \\ 
\midrule
\midrule
\multirow{2}{*}{\ours{-L/32}}  & \multirow{2}{*}{2,652} & \ours{-L/14} &   872 & 3,524 & 83.7 \\
  &  & \ours{-L/8} & 3,486 & 6,138 & 84.2 \\
 \midrule
  \ours{-L/14}  &  8,947 & \ours{-L/14} &   872  &  9,819 & 83.9 \\
\midrule
 \ours{-L/8}  &  35,511 & \ours{-L/8} & 3,486  & 38,997 &  85.1 \\

 \bottomrule
\end{tabular}%
\label{tab:compute_efficient_scaling}
\end{table*}

\textbf{Results and analysis.} 
(1) When fine-tuning with \ours{}-L/14 and using \ours{}-L/32 for pre-training on ImageNet-21K, this approach consumes about 35\% of the TPU v3 core-hours required by \ours{}-L/14, yet achieves a promising 83.7\% top-1 accuracy on ImageNet-1K, comparable to the 83.9\% achieved by \ours{}-L/14;
(2) When fine-tuning with \ours{}-L/8 and using \ours{}-L/32 for pre-training, this approach consumes just 15\% of the training time required by \ours{}-L/8, yet it still achieves a promising 84.2\% top-1 accuracy on ImageNet-1K, compared to 85.1\% when using the \ours{}-L/8 model in the pre-training stage; (3) While the total computational cost of \ours{}-L/32 + \ours{}-L/8 is less than that of \ours{}-L/14 + \ours{}-L/14, the performance of the former is slightly better. 
In summary, we find that this strategy offers a valuable reference for efficiently scaling \ours{} models in the future.

\begin{table*}[t!]
\centering
\caption{\textbf{Object detection and fine-grained segmentation via MaskCut on COCO {val2017}~\citep{lin2014microsoft}}.  
We evaluate models of various scales and assess their average precision using COCO's official evaluation metric. 
Compared with existing models such as  \textsc{crate} and ViT, \ours{} model achieves a notable performance gain.
In addition, when scaling \ours{} from base to large, it also exhibits the benefit of scalability.}
\small
\setlength{\tabcolsep}{2pt}
\resizebox{0.95\textwidth}{!}{\begin{tabular}{@{}llcccccccc@{}}
\toprule
 &  & \multicolumn{3}{c}{Detection} &  \multicolumn{3}{c}{Segmentation} \\ 
Model & Train & AP$_{50} \uparrow $ & AP$_{75} \uparrow $ & AP $\uparrow$ & AP$_{50} \uparrow$ & AP$_{75} \uparrow $ & AP $\uparrow$ \\ 
\midrule
 \textsc{crate}-B/8 \cite{yu2024emergence} & Supervised & 2.9& 1.0 & 1.3 & 2.2 & 0.7 & 1.0 \\
ViT-B/8~\cite{yu2024emergence} & Supervised & 0.8 & 0.2 & 0.4 & 0.7 & 0.5 & 0.4 \\
\midrule
\ours{}-B/8 & Supervised & 3.5 & 1.1 & 1.5 & 2.2 & 1.0 & 1.1 \\
\ours{}-L/8 & Supervised & 4.0 & 1.7 & 2.0 & 2.7 & 1.1 & 1.4 \\
\bottomrule
\end{tabular}}
\label{tab: maskcut_main_result}
\vspace{-0.15in}
\end{table*}

\vspace{-0.1in}
\section{Discussion}
\label{sec:limitation}
\vspace{-0.05in}
\textbf{Limitations.} Although we have used some existing compute-efficient training methods (e.g., CLIPA~\cite{li2024inverse}) and have initiated an exploration into compute-efficient scaling strategies for white-box transformers in Section~\ref{sec:efficient_scaling}, this work still requires a relatively large amount of computational resources, which may not be easily accessible to many researchers. 

\textbf{Societal impact.} A possible broader implication of this research is the energy consumption needed to conduct the experiments in our scaling study. However, there is growing interest in developing white-box transformers for better interpretability and transparency across a wide range of tasks and domains, including image segmentation \cite{yu2023white}, self-supervised masked autoencoders \cite{pai2023masked}, and integrated sensing and communications \cite{zhang2024white}, etc. Moreover, our results on the scalability of white-box transformers could also shed light on scaling up a broader class of white-box deep neural networks, such as white-box ISTA networks and their variants \cite{gregor2010learning,sun2018supervised,chen2018theoretical,zarka2019deep,li2022revisiting},  designed via unrolled optimization. In summary, we believe that our findings and insights could be helpful for developing white-box transformers for a wide range of applications and tasks, benefiting a broad audience interested in building more interpretable and performant deep learning models and further amortizing the pre-training compute costs.

\vspace{-0.1in}
\section{Conclusion}
\vspace{-0.05in}
This paper provides the first exploration of training white-box transformer \textsc{crate} at scale for vision tasks. We introduce both principled architectural changes and improved training recipes to unleash the potential scalability of the \textsc{crate} type architectures. 
With these modifications, we successfully scale up the \ours{} model along both the dimensions of model size and data size, while preserving, in most cases even improving, the semantic interpretability of the learned white-box transformer models. We believe this work provides valuable insights into scaling up mathematically interpretable deep neural networks, not limited to transformer-like architectures.

\subsection*{Acknowledgement}
This work is supported by a gift from Open Philanthropy, TPU Research Cloud (TRC) program, and Google Cloud Research Credits program.

{
\bibliographystyle{plainnat}
\bibliography{reference}
}

\newpage

\appendix

\begin{center}
    \textbf{\Large Appendix}
\end{center}
\vspace{0.1in}

\section{Additional  Experiments and Details}\label{sec:appendix-details}
\subsection{Model configuration.}\label{sec:appendix-model-details}

We provide details about \ours{} model configurations in Table~\ref{tab:model_configs}.

\vspace{-0.15in}
\begin{table}[ht]
    \centering
  \caption{Model configurations for different sizes of \ours{}, parameter counts, and comparisons to \textsc{crate} models.   \(L\) is depth,  \(d\)   is the hidden size, and  \(K\)  is the number of heads.}
  \vspace{0.05in}
    \footnotesize
    \setlength{\tabcolsep}{13pt}  %
        \begin{tabular}{@{}lccccc@{}}
            \toprule
            \textbf{Model Size} & \(L\) & \(d\) & \(K\) & \ours{} \# Params & \textsc{crate} \# Params
            \\ 
            \midrule
            \midrule
            Tiny & 12 & 192 & 3 & 4.8M & 1.7M \\
            \midrule
            Small & 12 & 576 & 12 & 41.0M & 13.1M \\
            \midrule
            Base & 12 & 768 & 12 & 72.3M & 22.8M \\
            \midrule
            Large & 24 & 1024 & 16 & 253.8M & 77.6M \\
            \midrule
            Huge & 32 & 1280 & 16 & 526.8M & 159.8M \\
            \bottomrule
        \end{tabular}%

    \label{tab:model_configs}
\end{table}

\begin{table*}[htb!]
\caption{The comparison between CRATE-$\alpha$ and ViT. FLOPs and throughput are calculated based on an input size of 224x224 on an NVIDIA RTX A6000 graphics card.}
\vspace{-0.05in}
\centering
\small
\setlength{\tabcolsep}{6pt} 
\resizebox{0.95\textwidth}{!}{
\begin{tabular}{lcccccccc}
\toprule
Model & FLOPs (G) & \#Params (M) & Throughput & Model & FLOPs (G) & \#Params (M) & Throughput \\ 
\midrule
CRATE-$\alpha$-B/32 & 6.4 & 74.0 & 499 & ViT-B/32 & 4.4 & 88.2 & 706 \\
CRATE-$\alpha$-B/16 & 25.8 & 72.3 & 233 & ViT-B/16 & 17.6 & 86.5 & 375 \\
CRATE-$\alpha$-L/32 & 22.8 & 256.0 & 215 & ViT-L/32 & 15.4 & 306.5 & 329 \\
CRATE-$\alpha$-L/14 & 119.7 & 253.7 & 56 & ViT-L/14 & 81.1 & 304.1 & 85 \\
\bottomrule
\end{tabular}%
}
\label{tab:crate_vit_comparison}
\end{table*}

\subsection{Comparison of model structure with ViT.}\label{sec:comparison_vit}
We also compare \ours{} to ViT in terms of computational costs, number of parameters, and inference speed. These comparisons are summarized in Table~\ref{tab:crate_vit_comparison}, where \ours{} matches ViT's efficiency while achieving similar accuracy. With the same number of layers and embedding dimensions, \ours{} has fewer parameters than ViT, and its FLOPs/Throughput is slightly higher. 

To more accurately compare \ours{} and ViT with larger model sizes, we conduct experiments on \ours{-L/16} with an image resolution of 336, nearly matching the setup of ViT-L/16. Both models use a similar amount of FLOPs: 210G for \ours{-L/16} compared to 191G for ViT-L/16. The throughput, or images processed per second, is also comparable at 35.53 for our model versus 35.56 for ViT-L/16. The accuracy of \ours{-L/16} reach 84.6\%, closely approaching ViT’s 85.2\% under similar conditions. Meanwhile, combining the trend from Figure \ref{fig:ablation_component} (right) in the main paper, this narrowing performance gap from Base to Large model size suggests that \ours{} can nearly matche ViT’s performance in large-scale settings. Besides, \ours{} inherits the mathematical interpretability of the white-box models and can also achieve much better semantic interpretability evaluated by zero-shot segmentation.

\subsection{Training details of \ours{}-CLIPA models.} When employing the \ours{} architecture to replace the vision encoder in the CLIPA~\cite{li2024inverse} framework, we essentially follow the original CLIPA training recipe. The setup for the pre-training stage is presented in Table~\ref{tab:hyper_clipa}. During the fine-tuning stage, we made some modifications: the input image size is set to \(224 \times 224\), the warmup steps are set to 800, and the base learning rate is set to 4e-7. When calculating the loss, we use the classification token from the vision encoder as the image feature and the last token from the text encoder as the text feature.

To explore the performance ceiling, we also train a ViT-CLIPA model from scratch. Most of the hyperparameters remain the same as those in Table~\ref{tab:hyper_clipa}, but there are some modifications in the pre-training stage. The batch size is set to 65,536, and the text length is set to 8 to speed up training. As with the CLIPA setup, warm-up steps are set to 3,200. Additionally, we add color jitter and grayscale augmentation, and use global average pooling instead of the classification token. These modifications help stabilize training.

\begin{table}[h]
    \centering
    \begin{tabular}{c|c}
         Config &Value \\
         \midrule
         optimizer & AdamW \cite{loshchilov2017adamw} \\
         optimizer momentum & (0.9, 0.95) \\
         batch size & 32768 \\
         base lr   & 8e-6 \\
         minimal lr & 0 \\
        warm-up steps & 1600 \\
        schedule & cosine decay \cite{loshchilov2016warmup} \\
         weight decay & 0.2  \\
         random crop area & (40, 100) \\
         resize method & bi-linear \\
         temperature init & 1/0.07 \cite{openclip, li2022flip}
    \end{tabular}
    \vspace{0.1in}
    \caption{\textbf{Pre-training hyper-parameters for CLIPA.}}
    \label{tab:hyper_clipa}
    \vspace{-1.em}
\end{table}

\begin{figure*}[htb!]
     \centering
     \begin{subfigure}[b]{0.98\textwidth}
         \centering
    \includegraphics[width=\textwidth]{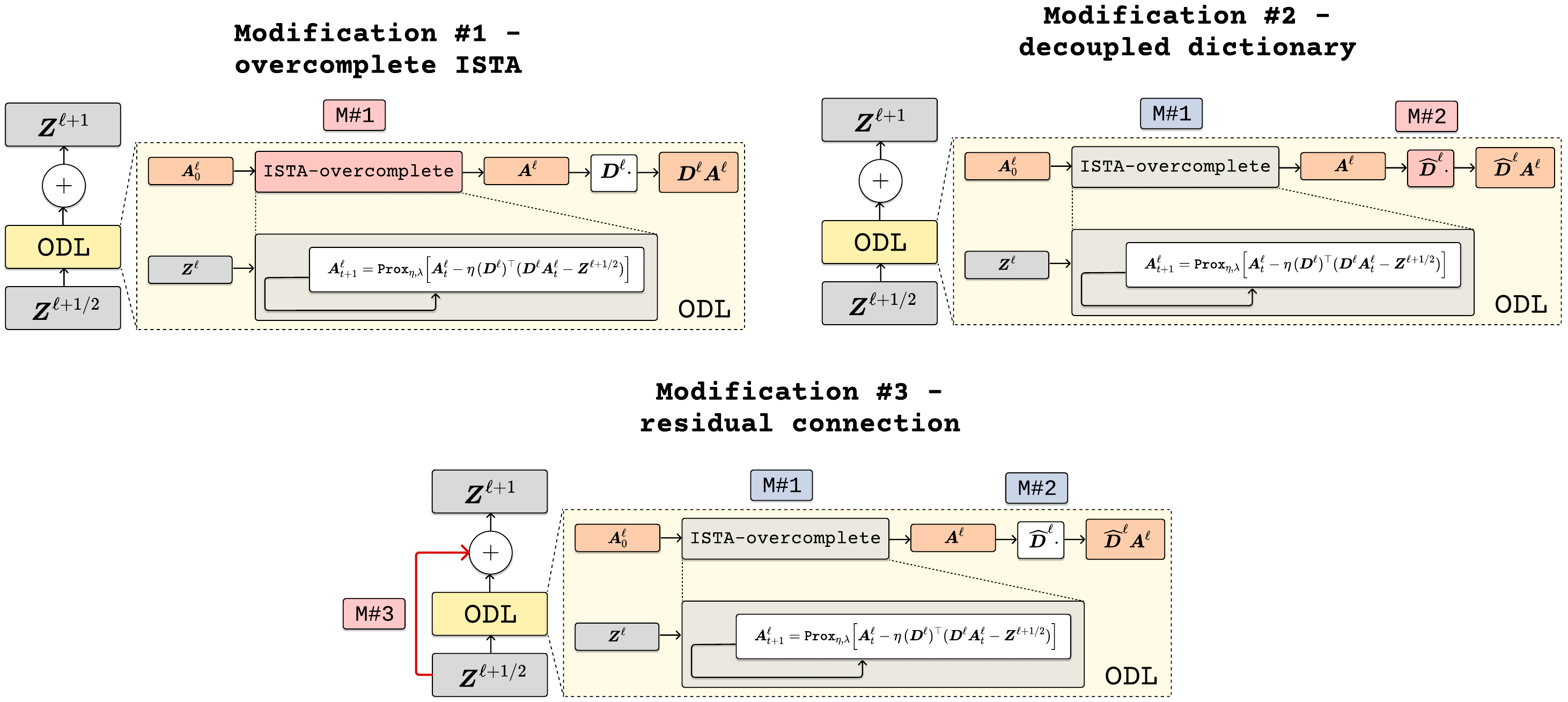}
     \end{subfigure}
     \vspace{0.5mm}
        \caption{One layer of the CRATE-$\alpha$ model architecture (with more details for the three modifications described in Section \ref{sec:crate-alpha}.}
        \label{fig:exp-rc-sparisty-small-new}
        \vspace{-0.2in}
\end{figure*}

\begin{figure}[ht]
    \centering
    \includegraphics[width=.99\textwidth]{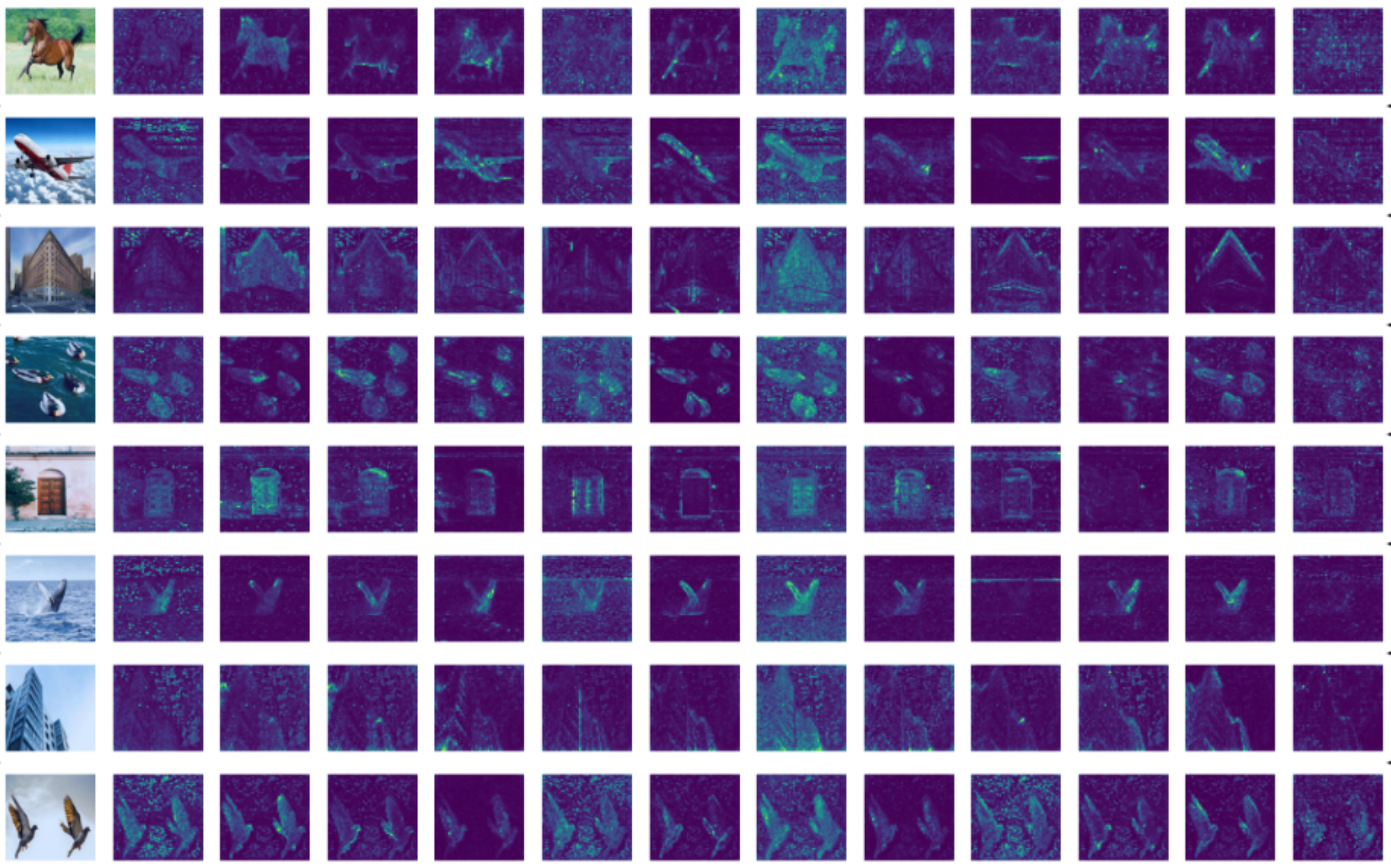}
    \caption{We visualize the self-attention maps of the \ours{} Base model using $8 \times 8$ patches trained using classification. Similar to the original CRATE \cite{yu2024emergence}, our model also demonstrates the capability to automatically capture the structural information of objects. For each row, the original image is displayed on the left, while the corresponding self-attention maps are shown on the right. The number of self-attention maps corresponds to the number of heads in the \ours{} model.
    }
    \vspace{-1.5em}
    \label{fig:vis_attention_crate_alpha}
\end{figure}

\newpage
\textbf{Visualization of self-attention maps of \ours{}.} We provide visualization of attention maps of \ours{} in Fig.~\ref{fig:vis_attention_crate_alpha}.

\begin{figure}[ht]
    \centering
    \includegraphics[width=.7\textwidth]{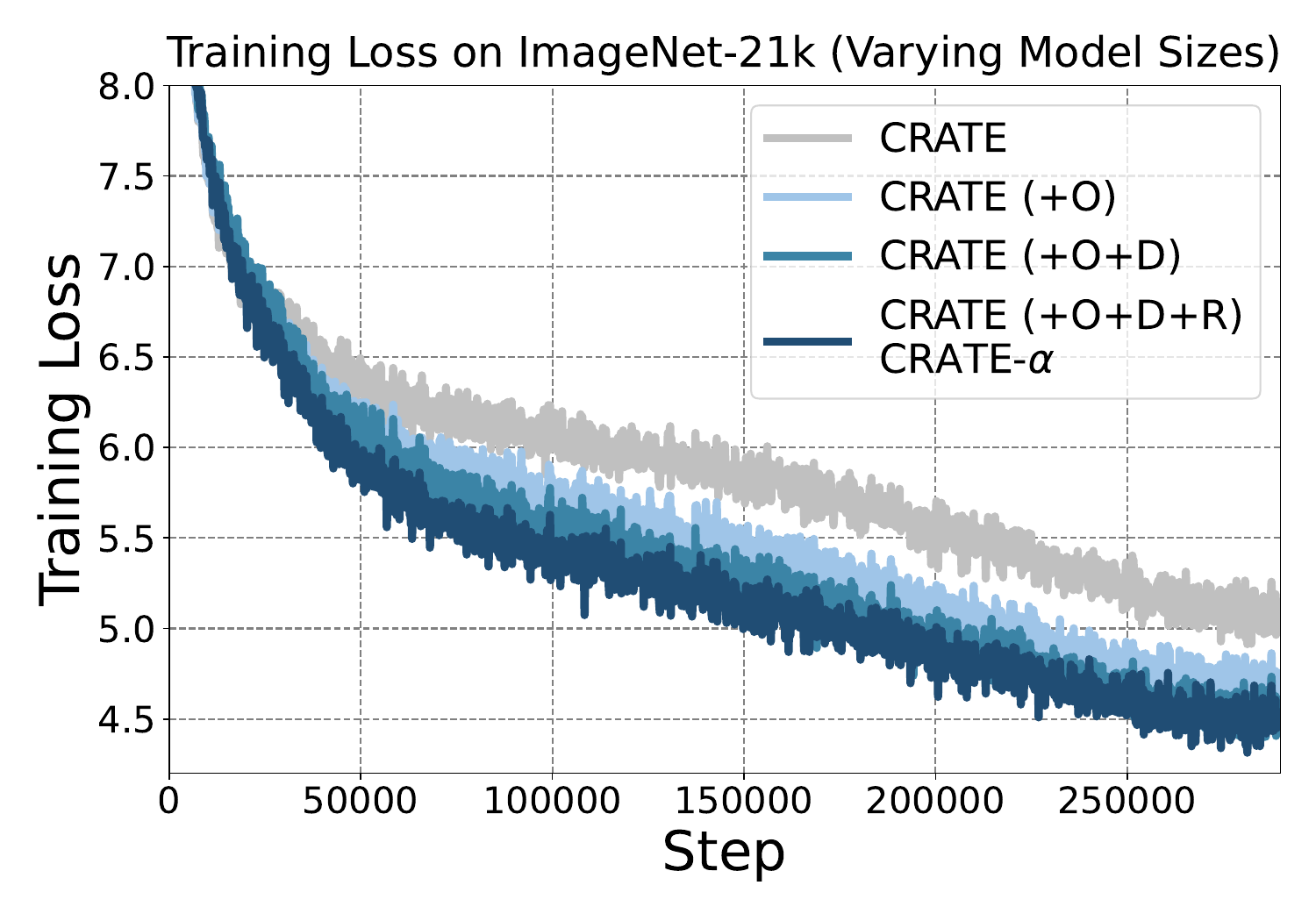}
    \caption{Training loss curves of different model architectures (mentioned in Fig. \ref{fig:ablation_component} in the main paper) on ImageNet-21K. The patch size is 32 for all four models shown in this figure. (+O: +overcomplete dictionary, +D: +decoupled dictionary, +R: +residual connection.)}
    \vspace{-1em}
    \label{fig:training_loss_curves_model_arch_ablation}
\end{figure}

\begin{figure}[ht]
    \centering
    \includegraphics[width=.7\textwidth]{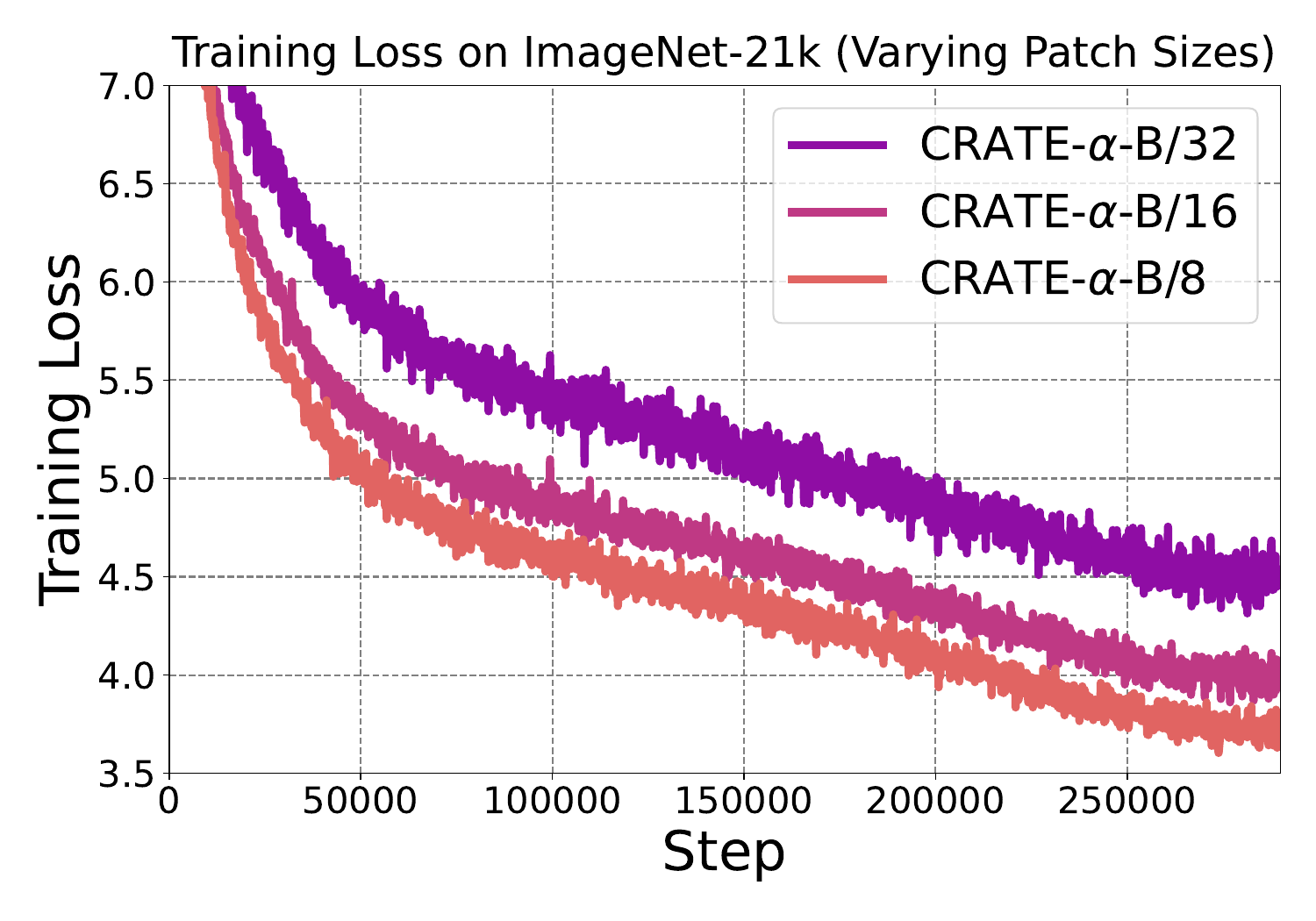}
    \caption{Comparing training loss curves across \ours{-Base} with different patch sizes.}
    \vspace{-1em}
    \label{fig:training_loss_curves_vary_patchsize_base}
\end{figure}

\begin{figure}[ht]
    \centering
    \includegraphics[width=.7\textwidth]{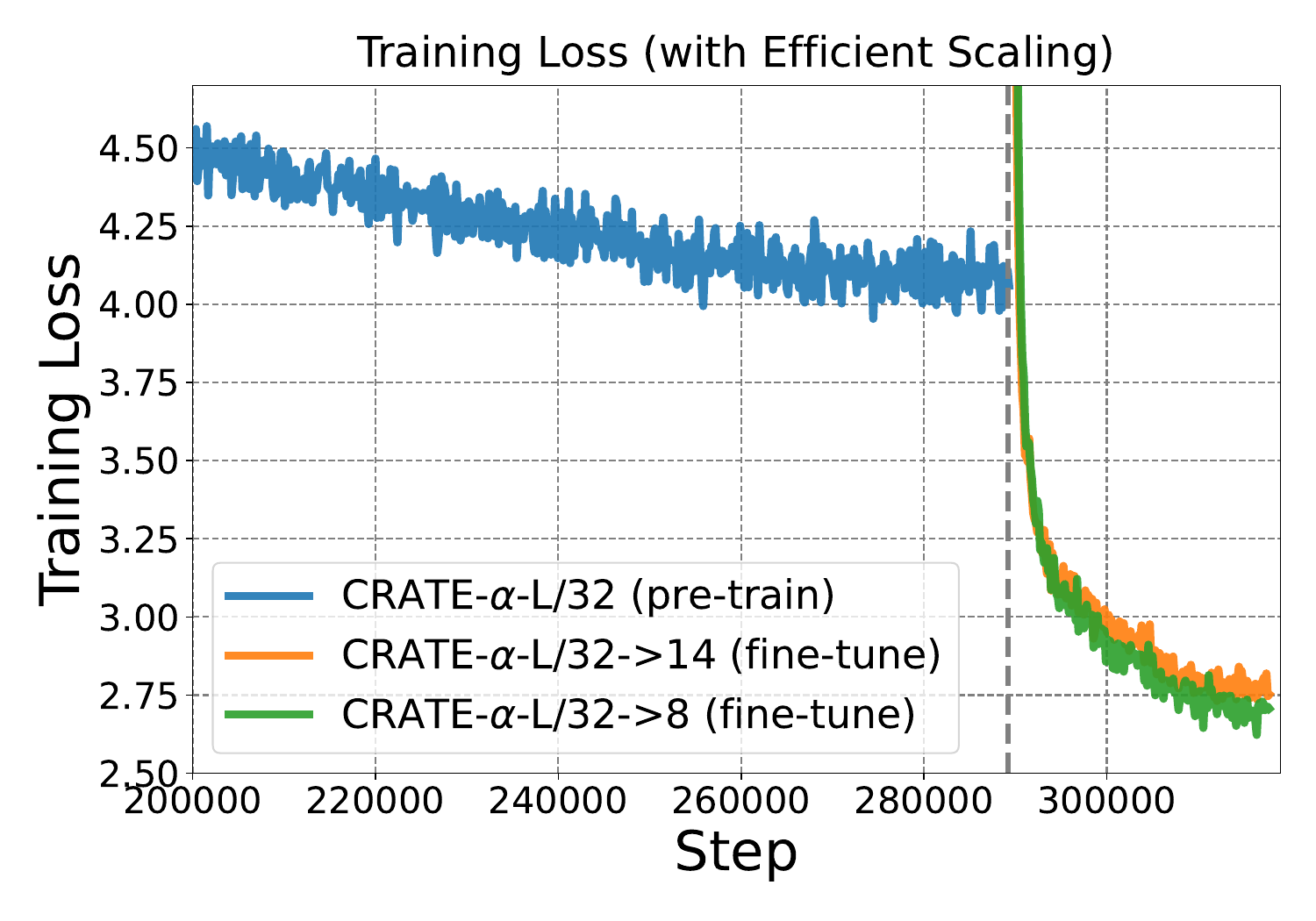}
    \caption{Comparing training loss curves when using the efficient scaling strategy. The blue curve corresponds to the \ours{-Large/32} model (in the pre-training stage). After pre-training the \ours{-Lage/32}, we further fine-tune it with smaller patch sizes  (longer token length), including patch size 14 (orange curve) and patch 8 (green curve).}
    \vspace{-1em}
    \label{fig:training_loss_curves_efficient_scaling}
\end{figure}

\vspace{0.1in}
\textbf{Visualization of loss curves.} 
We visualize the training loss curves of the four models, including \textsc{crate} and its three variants, in Fig.~\ref{fig:training_loss_curves_model_arch_ablation}. 
We visualize the training loss curves of \ours{-Base} with different patch sizes in Fig.~\ref{fig:training_loss_curves_vary_patchsize_base}. 
In Fig.~\ref{fig:training_loss_curves_efficient_scaling}, we also visualize the training loss curves of models trained with efficient scaling strategy described in Section \ref{sec:efficient_scaling} in the main paper.


\end{document}